\pdfoutput=1
\documentclass[11pt]{article}

\usepackage[]{emnlp2021}

\usepackage{times}
\usepackage{latexsym}

\usepackage[T1]{fontenc}
\usepackage[utf8]{inputenc}

\usepackage{microtype}

\usepackage{tabularx, booktabs}
\usepackage{multirow}
\usepackage[inline]{enumitem}
\usepackage{wrapfig}
\usepackage{array}
\usepackage{diagbox}
\usepackage{xspace}
\usepackage{url}
\usepackage{graphicx}
\usepackage{subfigure}
\usepackage[noend,ruled]{algorithm2e}
\usepackage{amsmath}
\usepackage{amssymb,mathtools,bm,etoolbox}
\title{``Average'' Approximates ``First Principal Component''? An Empirical Analysis on Representations from Neural Language Models}

\author{
Zihan Wang $^1\;\;\;\;\;$ Chengyu Dong $^1\;\;\;\;\;$ Jingbo Shang $^{*, 1,2}$ \\
\small $^1$ Department of Computer Science and Engineering, University of California San Diego, CA, USA \\
\small $^2$ Hal\i c\i o\u glu Data Science Institute, University of California San Diego, CA, USA \\
\small \texttt{\{ziw224, cdong, jshang\}@ucsd.edu}
\thanks{$^*$Jingbo Shang is the corresponding author.}
}

\date{}

\newcommand{\smallsection}[1]{\noindent{\textbf{#1.}}}

\begin{document}
\maketitle

\begin{abstract}
Contextualized representations based on neural language models have furthered the state of the art in various NLP tasks.
Despite its great success, the nature of such representations remains a mystery. 
In this paper, we present an empirical property of these representations---\emph{``average'' $\approx$ ``first principal component''}.
Specifically, experiments show that the average of these representations shares almost the same direction as the first principal component of the matrix whose columns are
these representations.
We believe this explains why the average representation is always a simple yet strong baseline. 
Our further examinations show that this property also holds in more challenging scenarios, for example, when the representations are from a model right after its random initialization.
Therefore, we conjecture that this property is intrinsic to the distribution of representations and not necessarily related to the input structure.
We realize that these representations empirically follow a normal distribution for each dimension, and by assuming this is true, we demonstrate that the empirical property can be in fact derived mathematically.

\end{abstract}

\section{Introduction}
A large variety of state-of-the-art methods in NLP tasks nowadays are built upon contextualized representations from pre-trained neural language models, such as ELMo~\cite{DBLP:conf/naacl/PetersNIGCLZ18}, BERT~\cite{DBLP:conf/naacl/DevlinCLT19}, and XLNet~\cite{DBLP:conf/nips/YangDYCSL19}.
Despite the great success, we lack understandings about the nature of such representations. 
For example, \citet{DBLP:conf/acl/AharoniG20} have shown that averaging BERT representations in a sentence can preserve its domain information. 
However, to our best knowledge, there is no analysis on what leads to the power of averaging representations.

\begin{figure}
    \centering
    \includegraphics[width=1\linewidth]{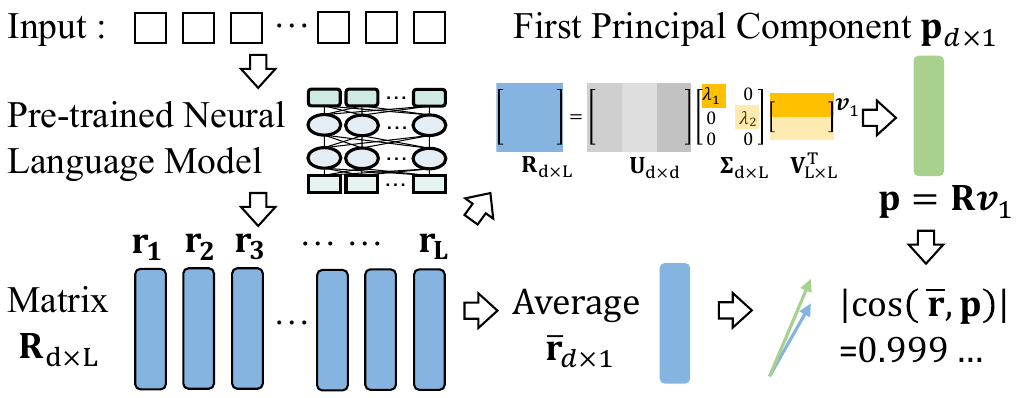}
    \vspace{-6mm}
    \caption{
    Visualization of our discovered empirical property: \emph{``average'' $\approx$ ``first principal component''}. 
    }%
    \label{fig:pipeline}%
\end{figure}

In this work, we present an empirical property of these representations, \emph{``average'' $\approx$ ``first principal component''}.
As shown in Figure~\ref{fig:pipeline}, given a sequence of $L$ tokens, one can construct a $d\times L$ matrix $\mathbf{R}$ using each $d$-dimensional representation $\mathbf{r_i}$ of the $i$-th token as a column.
There are two popular ways to project this matrix into a single $d$-dimensional vector: (1) \emph{average} and (2) \emph{first principal component}. 
Formally, the average $\overline{\mathbf{r}}$ is a $d$-dimensional vector where $\overline{\mathbf{r}} = \sum_{i = 1}^L \mathbf{r_i} / L$.
The first principal component $\mathbf{p}$ is a $d$-dimensional vector whose direction maximizes the variance of the (mean-shifted) $L$ representations. 
Then, the property can be written as $|\mbox{cos}(\overline{\mathbf{r}}, \mathbf{p})| \approx 1$. 
This absolute value is more than 0.999 in our experiments.

\begin{table}[t]
\centering
\renewcommand\tabcolsep{2pt}
\renewcommand\arraystretch{.85}
\caption{Average and minimum absolute cosine similarity of last layer representations between $\overline{\mathbf{r}}$ and $\mathbf{p}$ from 4,000 tests. 
As a reference, $\mathbf{r_i}$ drawn from a uniformly random distribution would lead to Average of .0149. 
}
\vspace{-3mm}
\label{tbl:three_datasets}
\small
\begin{tabular}{l c c c c c c} 
\toprule
\multirow{2}{*}{\textbf{Model}} & \multicolumn{2}{c}{
\textbf{AG's news}} & \multicolumn{2}{c}{\textbf{KP20k}} & \multicolumn{2}{c}{\textbf{Dbpedia}}\\
& Average & Min & Average & Min & Average & Min\\
% \textbf{Model} & \textbf{Average} & \textbf{Min} \\
\midrule
BERT & .9994 & .9908 & .9995 & .9958 & .9988 & .9845 \\
RoBERTa & .9989 & .9984 & .9990 & .9982 & .9987 & .9980 \\
XLNet & .9990 & .9874 & .9991 & .9932 & .9994 & .9856 \\
ELMo & .9957 & .9681 & .9985 & .9666 & .9949 & .9355 \\
\midrule
Word2vec & .9590 & .8506 & .9647 & .8907 & .9530 & .8474 \\
Glove & .9639 & .5014 & .9839 & .6369 & .9697 & .6088 \\
\bottomrule
\end{tabular}
\vspace{-3mm}
\end{table}

% \begin{table}[t]
% \centering
% \caption{Average and minimum absolute cosine similarity between $\overline{\mathbf{r}}$ and $\mathbf{p}$ from 4,000 tests from AG's news. }
% \vspace{-3mm}
% \label{tbl:main}
% \small
% \setlength{\tabcolsep}{3pt}
% \scalebox{0.9}{
% \begin{tabular}{l c c c c c c} 
% \toprule
% \multirow{3}{*}{\textbf{Model}} & \multicolumn{2}{c}{
% \textbf{Same Sentence}} & \multicolumn{2}{c}{\textbf{Same Sentence}} & \multicolumn{2}{c}{\textbf{Random Sentence}}\\
% & \multicolumn{2}{c}{
% \textbf{+ Last Layer}} & \multicolumn{2}{c}{\textbf{+ Random Layer}} & \multicolumn{2}{c}{\textbf{+ Last Layer}} \\
% \cmidrule(lr){2-3} \cmidrule(lr){4-5} \cmidrule(lr){6-7}
% & Average & Min & Average & Min & Average & Min\\
% % \textbf{Model} & \textbf{Average} & \textbf{Min} \\
% \midrule
% BERT & .9994 & .9908 & .9995 & .9989 & .9992 & .9930 \\
% RoBERTa & .9989 & .9984 & .9989 & .9984 & .9989 & .9985 \\
% GPT-2 & .9986 & .9781 & .9988 & .9984 & .9986 & .9954 \\
% XLNet & .9990 & .9874 & .9994 & .9981 & .9994 & .9960 \\
% ELMo & .9957 & .9681 & N/A & N/A & .9860 & .6836 \\
% \midrule
% Word2vec & .9590 & .8506 & N/A & N/A & .9405 & .8497 \\
% Glove & .9639 & .5014 & N/A & N/A & .9546 & .3102 \\
% \bottomrule
% \end{tabular}
% }
% \vspace{-3mm}
% \end{table}
\begin{table}[t]
\centering
\caption{Average and minimum absolute cosine similarity between $\overline{\mathbf{r}}$ and $\mathbf{p}$ from 4,000 tests from AG's news. }
\vspace{-3mm}
\label{tbl:main}
\small
\setlength{\tabcolsep}{3pt}
% \scalebox{0.9}{
\begin{tabular}{l c c c c} 
\toprule
\multirow{3}{*}{\textbf{Model}} & \multicolumn{2}{c}{
\textbf{Same Sentence}} & \multicolumn{2}{c}{\textbf{Random Sentence}}\\
\cmidrule(lr){2-3}  \cmidrule(lr){4-5}
& Average & Min & Average & Min\\
\midrule
BERT & .9994 & .9908 & .9992 & .9930 \\
RoBERTa & .9989 & .9984 & .9989 & .9985 \\
XLNet & .9990 & .9874 & .9994 & .9960 \\
ELMo & .9957 & .9681 & .9860 & .6836 \\
\midrule
Word2vec & .9590 & .8506 & .9405 & .8497 \\
Glove & .9639 & .5014 & .9546 & .3102 \\
\bottomrule
\end{tabular}
% }
\vspace{-3mm}
\end{table}

We examine the generality of this property and find it also holds in three more scenarios when every $\mathbf{r_i}$ is drawn from
(1) a fixed layer (not necessary the last layer) in a pre-trained neural language model,
(2) a fixed layer in a model right after random initialization without any training,
and (3) random token representations from all sentences encoded by a pre-trained model. 
Therefore, we conjecture that this property is intrinsic to the representations' distribution, which is related to the neural language model's architecture and parameters, and not necessarily related to the input structure. 
We realize that the empirical distribution of these representations is similar to a normal distribution on each dimension.
Assuming this is true, we show that the property can be in fact derived mathematically.

Our contributions are summarized as follow.
\begin{itemize}[leftmargin=*,nosep]
    \item We discover a common, insightful property of several pre-trained neural language models---\emph{``average'' $\approx$ ``first principal component''}. To some extent, this explains why the average representation is always a simple yet strong baseline. 
    \item We verify the generality of this property by obtaining representations from a random mixture of layers and sentences and also using randomly initialized models instead of pre-trained ones.
    \item We show that representations from language models empirically follow a per-dimension normal distribution that leads to the property.
\end{itemize}
\smallsection{Reproducibility}
We will release code to reproduce experiments on Github\footnote{\url{https://github.com/ZihanWangKi/AverageApproxFirstPC}}.

\section{Experimental Settings}
\smallsection{Dataset}
We random sample 4,000 sentences each from three different datasets on three different domains: AG's news corpus~\cite{DBLP:conf/nips/ZhangZL15}, KP20k Computer Science papers~\cite{DBLP:conf/acl/MengZHHBC17}, and DBpedia~\cite{DBLP:conf/nips/ZhangZL15}.

\smallsection{Pre-trained Neural Language Models}
We experiment on four well-known language models: (1) BERT~\cite{DBLP:conf/naacl/DevlinCLT19}, (2) RoBERTa~\cite{DBLP:journals/corr/abs-1907-11692}, (3) XLNet~\cite{DBLP:conf/nips/YangDYCSL19}, and (4) ELMo~\cite{DBLP:conf/naacl/PetersNIGCLZ18}.
For the first four transformer-based models, we use the base (and cased, if available) version from the HuggingFace's~\cite{DBLP:journals/corr/abs-1910-03771} implementation. 
For ELMo, we follow the AllenNLP toolbox~\cite{DBLP:journals/corr/abs-1803-07640}.

\smallsection{Word Embedding Models}
We include experiments on word embeddings Word2vec~\cite{DBLP:conf/nips/MikolovSCCD13} and Glove~\cite{DBLP:conf/emnlp/PenningtonSM14} learned on Wikipedia~\cite{DBLP:conf/nodalida/FaresKOV17}.

\section{The Property: ``Average'' $\approx$ ``First Principal Component''}
In most applications, each representation $\mathbf{r_i}$ in $\mathbf{R}$ comes from the tokens within the \textbf{\emph{Same Sentence}} and the last layer of a pre-trained neural language model. 
Following this setting, we conduct 4,000 tests each on three datasets and summarize the results in Table~\ref{tbl:three_datasets}.
One can easily see that the \emph{average} and \emph{min} absolute cosine similarities are very close to 1 for all pre-trained neural language models.
The word embeddings satisfy the property on average, but not for some outlier sentences~\footnote{We do not find obvious patterns like length or repeated words in the outlier sentences.}. 
Given that uniformly random representations have near-zero average and min absolute cosine similarity values, we conclude that this is a special property for the language model generated representations.
To some extent, it explains the effectiveness of the average last-layer representation based on a language model, which has been widely adopted and observed in the literature.

\begin{figure}[t]
    \centering
    \subfigure[Pre-trained]{
        \centering
        \includegraphics[width=0.46\linewidth]{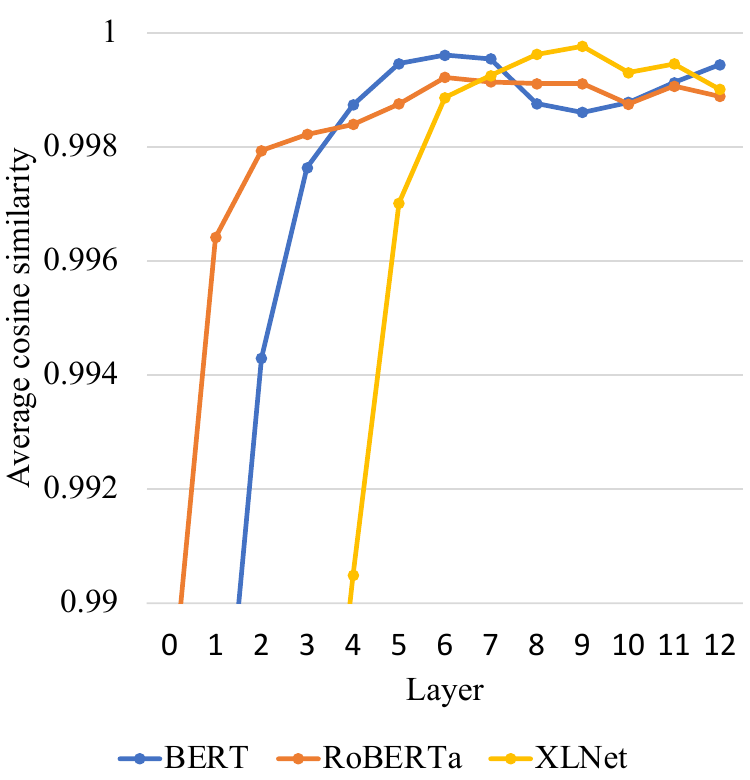}
    }
    \subfigure[Randomly Initialized]{
        \centering
        \includegraphics[width=0.46\linewidth]{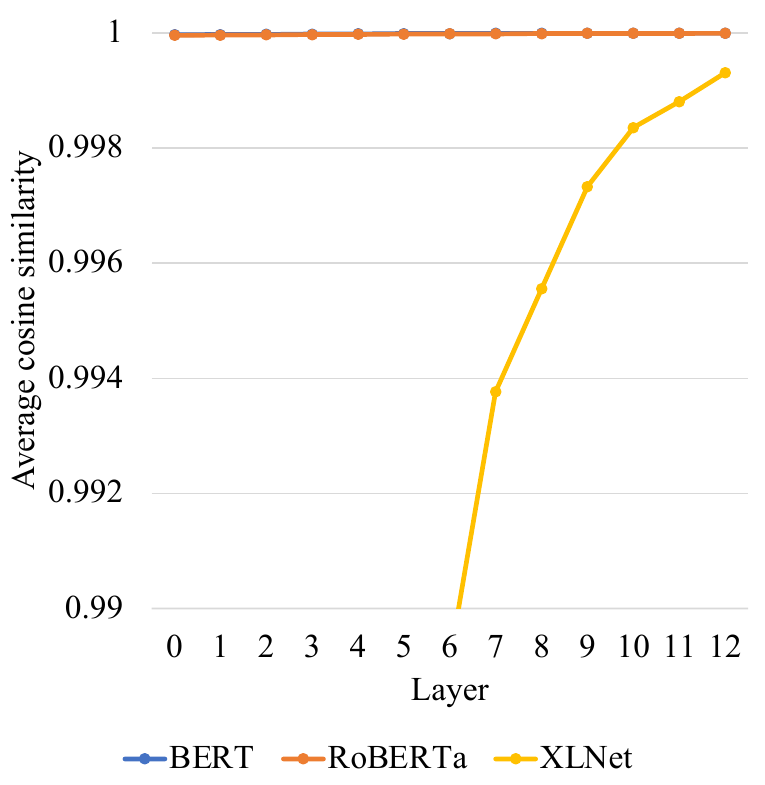}
    }
    \vspace{-3mm}
    \caption{Average cosine similarity for different layers.}
    \label{fig:layer_sim}
    \vspace{-3mm}
\end{figure}
\section{Generality Tests of the Property}\label{sec:generality}

\smallsection{Different Layers}
To evaluate our discovered property's generality, we first investigate if this property only holds for the last-layer representations.
For the four transformer-based language models, there are 13 possible layers (i.e., one after lookup table and 12 after encoder/decoder layers) to retrieve representations for tokens.
Therefore, we test the property based on representations from each layer and plot the average absolute cosine similarities in Figure~\ref{fig:layer_sim}. 
One can see that the property holds for the last few layers in all four models.
% Interestingly, GPT-2 has this property even for its representations from the lookup table.

\smallsection{Random Initialized Models}
We repeat the same test for randomly initialized models, i.e., not (pre-)trained at all. 
The results are in Figure~\ref{fig:layer_sim}. 
Again, we can see that the property holds for the last few layers in all four models.
% For both BERT and RoBERTa, the property holds right after the lookup table.

% \smallsection{Random Layer}
% We know that the property consistently holds for the last few layers from previous discussions.
% Here, we further explore to mix representations from different layers and test the property again.
% Specifically, for each token in a sentence, we randomly select a layer from 7 to 12 and use that layer's representation as $\mathbf{r_i}$. 
% The results are in the \textit{Same Sentence + Random Layer} section\footnote{Since the ELMo and word embedding models only have limited layers, they are not tested.} in Table~\ref{tbl:main}.
% Even when the representations are from random layers, the property still holds.

\smallsection{Random Sentence}
Finally, we explore the case when the representations can even come from different sentences. 
Specifically, we shuffle all the last-layer token representations of the 4,000 sentences and re-group them into 4,000 random lists of representations. 
With a high probability, each token representation in a list is generated independently of other tokens from the same list. 
We show the results in \textit{Random Sentences} section in Table~\ref{tbl:main}. 
Surprisingly, even with ``unrelated'' token representations, the property still holds well. 
\section{Analysis}
In this section, we attempt to answer what could be a reason that the language models show this property. 
From Section~\ref{sec:generality}, we know that the property also holds for randomly initialized models. 
Such models know nothing about natural languages.
Therefore, it is reasonable to believe that this property is intrinsic to the models and related to the distribution of these representations.

\subsection{Representation Distribution Analysis: BERT as a Case Study}
We show that each dimension of BERT representations likely follows a normal distribution.

\begin{figure}[t]
    \centering
    \includegraphics[width=0.75\linewidth]{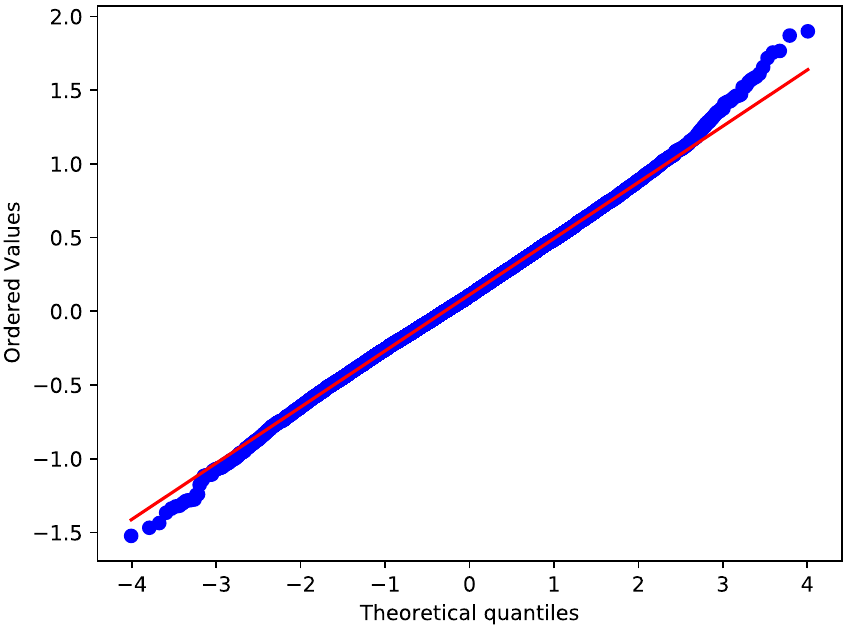}
    \vspace{-3mm}
    \caption{Q-Q plot of the 1st dimension of BERT representations against a normal distribution. We sampled 10\% of representations to reduce the figure size.}
    \label{fig:qqplot}
    \vspace{-3mm}
\end{figure}

From Figure~\ref{fig:qqplot}, we can see that the quantiles match with a normal distribution almost perfectly through a Q-Q plot~\cite{10.2307/2334448} on the first dimension. 
We have checked another ten random dimensions and their quantiles all match well (see Appendix). 

We also compare the skewness and kurtosis of a standard normal distribution and the empirical distribution of standardized representation values in each dimension.
Let $\mathbf{s_j}$ be the vector that contains values of dimension $j$ in the representations. Specifically, consider the representation matrix $\mathbf{R'}$ for all $D = 224,970$ representations over the 4,000 sentences. The rows of $\mathbf{R'}$ correspond to $\mathbf{s_j}$. The standardized vector $\widetilde{\mathbf{s_j}}$ of $\mathbf{s_j}$ is defined as $\widetilde{s_{ji}} = \frac{s_{ji} - \hat{\mu}_j}{\hat{\sigma}_j}$, where $\hat{\mu}_j = \frac{\sum_{i=1}^{D}s_{ji}}{D}$ and $\hat{\sigma_j} = \sqrt{\frac{\sum_{i=1}^D (s_{ji} - \hat{\mu}_j)^2}{D}}$. 
For each dimension $j, 1 \leq j \leq d$, one can obtain an empirical distribution from  $\widetilde{\mathbf{s_j}}$.
From Table~\ref{tbl:dist_approx}, the third moment matches with a standard normal distribution well, while the fourth moment is a bit off. Further, we examine the the off diagonal terms in the $d \times d$ covariance matrix of the representations, which has a mean of 0.0101 and a standard deviation of 0.0116. When compared with a mean of 0.1747 of the diagonal terms, this is very small. 
Therefore, we conjecture that each dimension of BERT's representation can be treated approximately like an independent normal distribution. We note that we do not perform normality tests due to the large dataset size (i.e., over 200,000 representations), since even a minor shift away from the normal distribution can make statistical tests reject the null hypothesis. % like Shapiro–Wilk test~\cite{shapiro1965analysis}, Kolmogorov–Smirnov test~\cite{massey1951kolmogorov}, or Pearson's chi-squared test~\cite{plackett1983karl} reject the null hypothesis.

\begin{table}[t]
\centering
\caption{$N(0, 1)$ vs. the Distribution of Normalized BERT Representations.
% $\Tilde{\mathbf{r}}$ is the normalized vector of $\mathbf{r}$. 
For empirical values, we show Avg($\pm$Std) over 768 dimensions.
% \jingbo{Why not use Avg($\pm$std)?}
}
\vspace{-3mm}
\label{tbl:dist_approx}
\small
\begin{tabular}{l c c} 
\toprule
% & \textbf{Normal} & \textbf{BERT} \\
 & $\sim N(0, 1)$ & $\sim \mbox{Distribution}(\widetilde{\mathbf{s_j}}$) \\
%  & \textbf{Mean} & \textbf{Min} \\
\midrule
% $\mathbb{E}[z]$   & 0 & 0.0000($\pm{0.0000}$) \\
% $\mathbb{E}[z^2]$ & 1 & 1.0000($\pm{0.0000}$) \\
Skewness ($\mathbb{E}[z^3]$) & 0 & 0.0062($\pm{0.5884}$) \\
Kurtosis ($\mathbb{E}[z^4]$) & 3 & 3.9629($\pm{3.3821}$) \\
\bottomrule
\end{tabular}
\end{table}

In the rest of this section, we assume representations are sampled from $d$ normal distributions, i.e., each dimension follows a distribution $N(\mu_j, \sigma_j^2)$. 

\subsection{Fitted distributions satisfy the property}
\begin{table}[t]
\centering
\caption{Property testing results of representations following $d$ normal distributions with $\mu_j$ and $\sigma_j$ sampled from certain uniform distributions. 4000 tests are done.}
\vspace{-3mm}
\label{tbl:normal_dist}
\small
\begin{tabular}{l c c} 
\toprule
$r_{ij} \sim N(\mu_j, \sigma_j^2)$ & Average & Min \\
\midrule
$\mu_j \sim \mathcal{U}[-1, 1], \sigma_j \sim \mathcal{U}[0, 1]$ & .9986 & .9939 \\
$\mu_j \sim \mathcal{U}[-1, 1], \sigma_j \sim \mathcal{U}[0, 10]$ & .1475 & .0000 \\
$\mu_j \sim \mathcal{U}[3, 5], \sigma_j \sim \mathcal{U}[0, 1]$ & .1490 & .1463 \\
$\mu_j = 0, \sigma_j \sim \mathcal{U}[0, 1]$ & .1587 & .0000 \\
\bottomrule
\end{tabular}
\vspace{-5mm}
\end{table}

% \begin{table}[t]
% \centering
% \caption{Multivariate normal distributed representations with parameters estimated from neural language models, or from certain distributions.}
% %  \jingbo{this caption could be improved.}
% \vspace{-3mm}
% \label{tbl:normal_dist}
% \small
% \begin{tabular}{l c c} 
% \toprule

% Model & Average & Min\\
% \midrule
% BERT & .9995 & .9978\\
% RoBERTa & .9989 & .9982 \\
% GPT & .9988 & .9982\\
% XLNet & .9994 & .9977 \\
% ELMo & .9987 & .9947\\
% % \midrule
% % Word2vec & .9979 & .9823 \\
% % Glove & .9942 & .9515 \\
% \midrule
% $\mu_j \sim \mathcal{U}[-1, 1], \sigma_j \sim \mathcal{U}[0, 1]$ & .9986 & .9939 \\
% $\mu_j \sim \mathcal{U}[-1, 1], \sigma_j \sim \mathcal{U}[0, 10]$ & .1475 & .0000 \\
% $\mu_j \sim \mathcal{U}[3, 5], \sigma_j \sim \mathcal{U}[0, 1]$ & .1490 & .1463 \\
% $\mu_j = 0, \sigma_j \sim \mathcal{U}[0, 1]$ & .1587 & 0 \\
% \bottomrule
% \end{tabular}
% \vspace{-3mm}
% \end{table}

We verify the property on generated representations following the distribution.
When the parameters $\mu_j, \sigma_j$ are estimated from representations from language models, the property holds (see Appendix). We can also randomly sample the parameters from pre-defined distributions, as shown in Table~\ref{tbl:normal_dist}. The results on pre-defined distributions tell us: (1) the average of all $\mu_j$ should be 0, (2) not all of $\mu_j$ should be exactly 0, and (3) the variance should not be too large in magnitude compared to the mean. 

In the following analysis, we additionally restrict that all representations have a sum of value to 0, i.e. $\sum_{j=1}^d r_{ij} = 0,$ for all representations $\mathbf{r_i}$. %This is mainly for not worrying about mean-shifting the $\mathbf{R}$ matrix in the PCA algorithm.
This is mainly for the simplicity of the covariance matrix computation, as the PCA algorithm will first mean-shift the R matrix.

\subsection{Covariance Matrix $\mathbf{C}$ of Normally Distributed Representations}
% Given that each representation in $\mathbf{R}$ sums to 0, we
We define the L-by-L covariance matrix $\mathbf{C} = \mathbf{R}^{\intercal}\mathbf{R}$.
Its L-by-1 eigenvector $\mathbf{w}$ corresponding to the largest eigenvalue can be used to get the first principal component, i.e., $\mathbf{p} = \mathbf{R} \mathbf{w}$.

We show that if the representations follow a per-dimension normal distribution, $\mathbf{C}$ will follow a special shape---by expectation, its diagonals and off-diagonals will be the same positive value, respectively. We theoretically derive the mean and standard deviation of the entries based on $\mu_j$ and $\sigma_j$ (derivations are available in Appendix), empirically estimate their values, and put them in Table~\ref{tbl:mean_and_std}.
It is clear that the standard deviation is smaller than the mean in magnitudes, confirming the special shape of $\mathbf{C}$.
Also, the theoretical and estimated values mostly match. 
The only significant difference is the standard deviation for diagonal entries, which is due to the difference on the fourth power statistics between the representations and the standard normal distribution as shown in Table~\ref{tbl:dist_approx}.

% We show that if the representations follow a per-dimension normal distribution, $\mathbf{C}$ will follow a special shape---by expectation, its diagonals and off-diagonals will be the same positive value, respectively. We theoretically derive the mean and standard deviation of the entries based on $\mu_j$ and $\sigma_j$ (derivations are available in Appendix):

% \vspace{-3mm}
% $$
% \scriptsize
% \begin{aligned}
% \mathbb{E}[C_{ii}] & = \frac{1}{d - 1} \sum_{k = 1}^d (\sigma_k^2 + \mu_k^2) \\
% Var[C_{ii}] & = \frac{1}{(d - 1)^2} \left(\sum_{k = 1}^d 2\sigma_k^4 + 4\mu_k^2\sigma_k^2\right) \\
% \mathbb{E}[C_{ij}] & = \frac{1}{d - 1} \sum_{k = 1}^d \mu_k^2 \\
% Var[C_{ij}] & = \frac{1}{(d - 1)^2} \left(\sum_{k = 1}^d \sigma_k^4 + 2\mu_k^2\sigma_k^2\right)
% \end{aligned}
% $$

% We also estimate their empirical values on BERT representations and put them in Table~\ref{tbl:mean_and_std}.

\subsection{This Special $\mathbf{C}$ $\rightarrow$ the Property}
If the diagonal entries of the covariance matrix $\mathbf{C}$ are $a > 0$, and all off-diagonal entries are $b > 0$, the eigenvector $\mathbf{w}$ corresponding to the largest eigenvalue will be a uniform vector. 
The Perron–Frobenius theorem~\cite{samelson1957} states that the (unique) largest eigenvalue $\lambda$ is bounded:

\vspace{-4mm}
\begin{equation}
\small
\begin{aligned}
\min_{i} \sum_{j=1}^L C_{ij} \leq \lambda \leq \max_{i} \sum_{j=1}^L C_{ij},
\end{aligned}
\end{equation}

\noindent which refer to the min and max row-sums in $\mathbf{C}$. 
Due to its special shape, all row-sums in $\mathbf{C}$ are around $a + b * (L - 1)$.
Therefore, the largest eigenvalue $\lambda_1 \approx a + b *  (L - 1)$. 
To obtain $\mathbf{w}$, one can solve $\mathbf{C} \mathbf{w} = \lambda_1 \mathbf{w}$.
Obviously, $\mathbf{w} = \mathbf{1}$ is a solution, where $\mathbf{1}$ is a vector of 1's of length $L$. 
As a result, the first principal component $\mathbf{p} = \mathbf{R}\mathbf{w}$  follows the same direction as the average.

\begin{table}[t]
\centering
\caption{Theoretical and Estimated Mean and Standard Deviation of the Values in the Covariance Matrix $\mathbf{C}$.}
\vspace{-3mm}
\label{tbl:mean_and_std}
\small
\begin{tabular}{l c c c c } 
\toprule
& \multicolumn{2}{c}{\textbf{Theoretical}} & \multicolumn{2}{c}{\textbf{Estimated}}\\
& Mean & Std & Mean & Std \\
\midrule
diagonal     & 0.2857 & 0.0350 & 0.2857 & 0.0710 \\
off-diagonal & 0.1100 & 0.0248 & 0.1100 & 0.0248 \\
\bottomrule
\end{tabular}
\vspace{-3mm}
\end{table}

\section{Related Work}
Simply averaging is a widely used, strong baseline to aggregate (contextualized) token representations~\cite{DBLP:conf/emnlp/Ethayarajh19,DBLP:conf/acl/AharoniG20,DBLP:conf/emnlp/ReimersG19,DBLP:conf/nips/ZhangZL15,DBLP:conf/acl/Taddy15,DBLP:conf/rep4nlp/YuLO18}. 
In this paper, we discover an empirical property of these representations (``average'' $\approx$ ``first principal component''), which can justify its effectiveness.

There are other attempts to analyze properties of language models.
\citet{DBLP:journals/corr/abs-1906-04341} analyze syntactic information that BERT's attention maps capture.
\citet{DBLP:conf/iclr/KWMR20} prune the causes for multilinguality of multilingual BERT.
\citet{DBLP:conf/emnlp/WangC20a} show that position information are learned differently in different language models. 
Different from these language-specific properties, we believe our newly discovered property relates more to the internal structure of neural language models. 

\section{Conclusion and Future Work}
This paper shows a common, insightful property of representations from neural language models---``average'' $\approx$ ``first principal component''. 
This property is general and holds in many challenging scenarios. 
After analyzing the BERT representations as a case study, we conjecture that these representations follow a normal distribution for each dimension, and this distribution leads to our discovered property. We believe that this work can shed light on future directions: (1) identifying the distributions that representations from language models follow, and (2) further implications or properties that representations have.

% \newpage
% \clearpage

\section*{Acknowledgements}
We thank anonymous reviewers and program chairs for their valuable and insightful feedback. 
The research was sponsored in part by National Science Foundation Convergence Accelerator under award OIA-2040727 as well as generous gifts from Google, Adobe, and Teradata.
Any opinions, findings, and conclusions or recommendations expressed herein are those of the authors and should not be interpreted as necessarily representing the views, either expressed or implied, of the U.S. Government. 
The U.S. Government is authorized to reproduce and distribute reprints for government purposes not withstanding any copyright annotation hereon.

% \section{Ethical Considerations}
% This paper provides an analysis on an existing phenomenon of language models and we do not anticipate any major ethical concerns.

\bibliography{anthology,custom}
\bibliographystyle{acl_natbib}

\newpage
\clearpage

\appendix
\section{Q-Q Plot of Ten Random Dimensions}
\begin{figure*}[h]
    \centering
    \subfigure[d = 29]{
        \centering
        \includegraphics[width=0.18\linewidth]{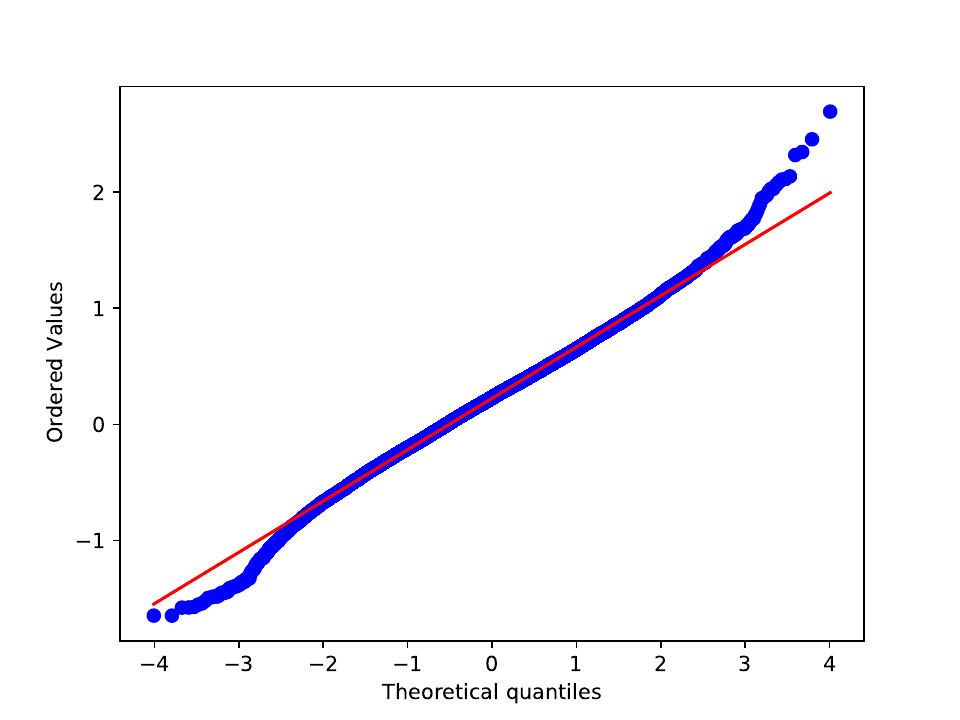}
    }
    \subfigure[d = 94]{
        \centering
        \includegraphics[width=0.18\linewidth]{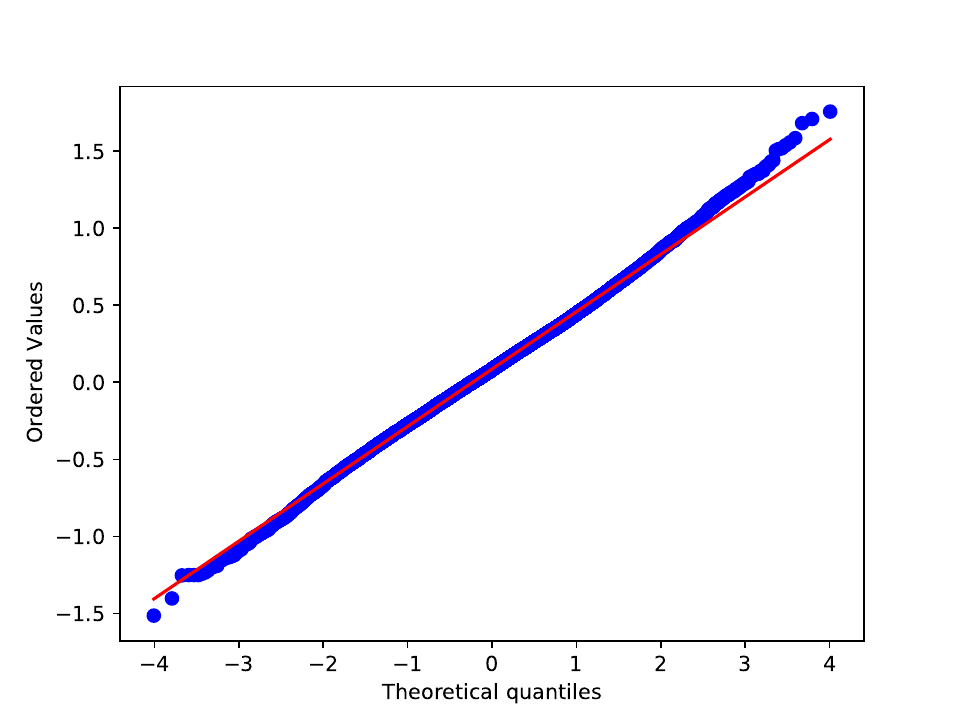}
    }
    \subfigure[d = 287]{
        \centering
        \includegraphics[width=0.18\linewidth]{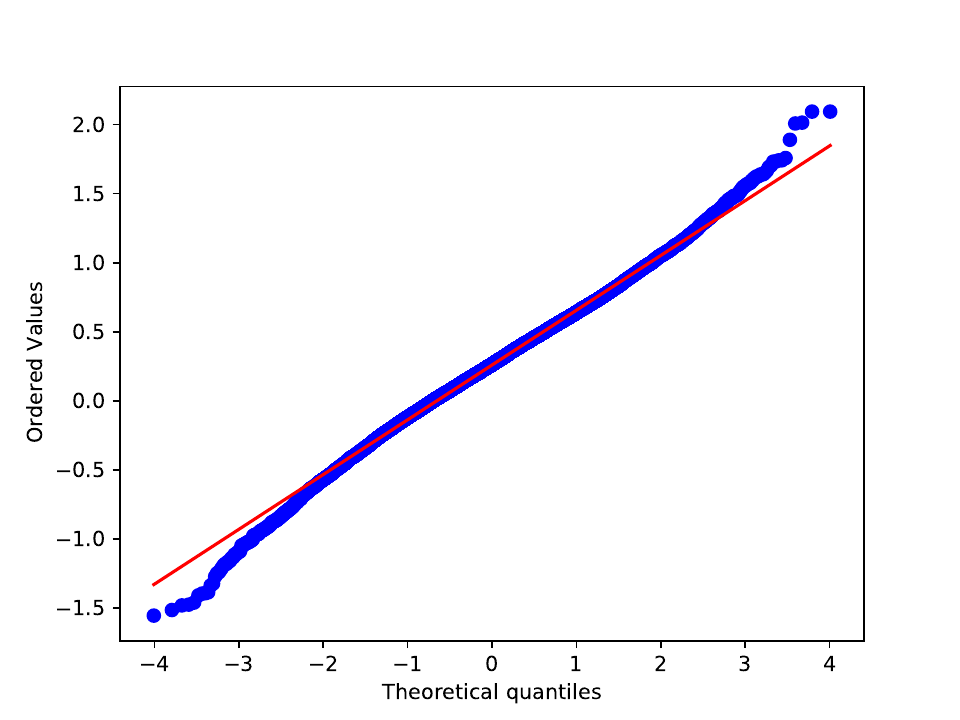}
    }
    \subfigure[d = 342]{
        \centering
        \includegraphics[width=0.18\linewidth]{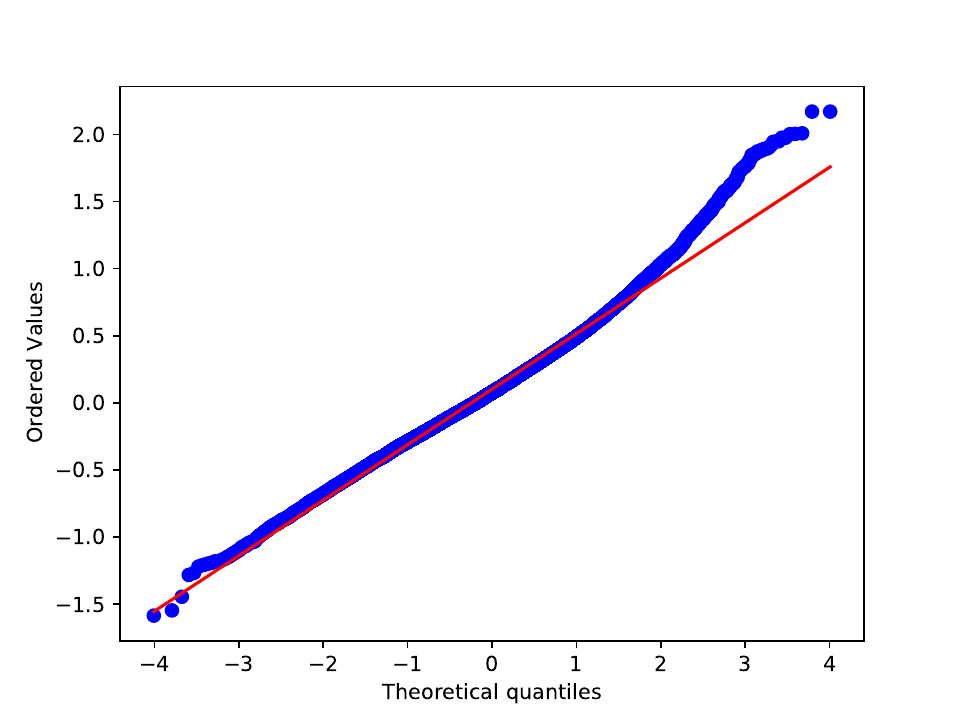}
    }
    \subfigure[d = 390]{
        \centering
        \includegraphics[width=0.18\linewidth]{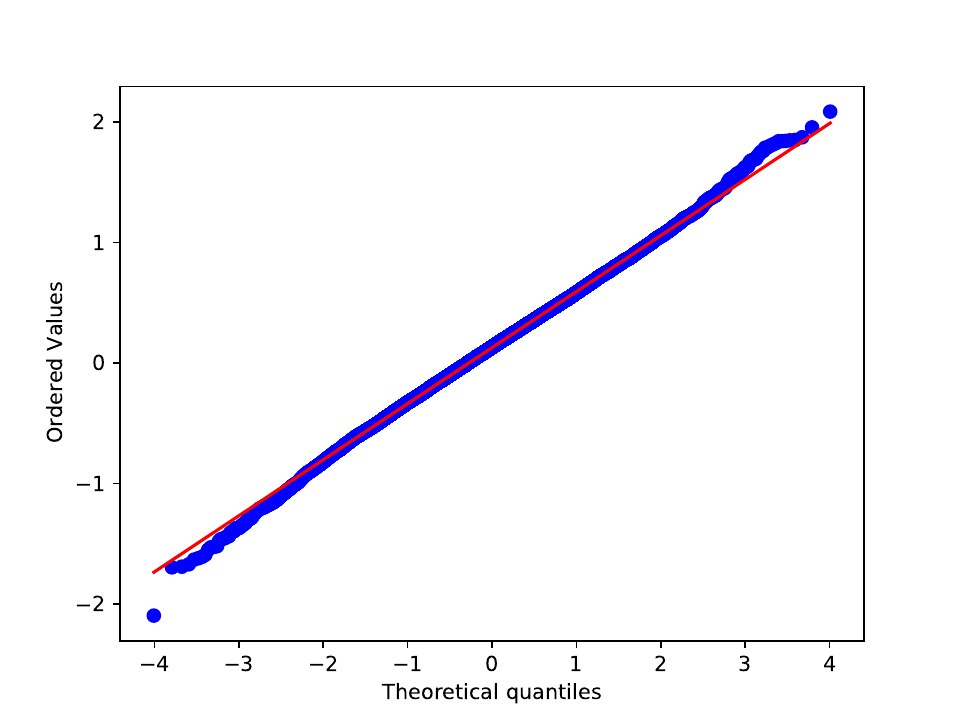}
    }
    \subfigure[d = 495]{
        \centering
        \includegraphics[width=0.18\linewidth]{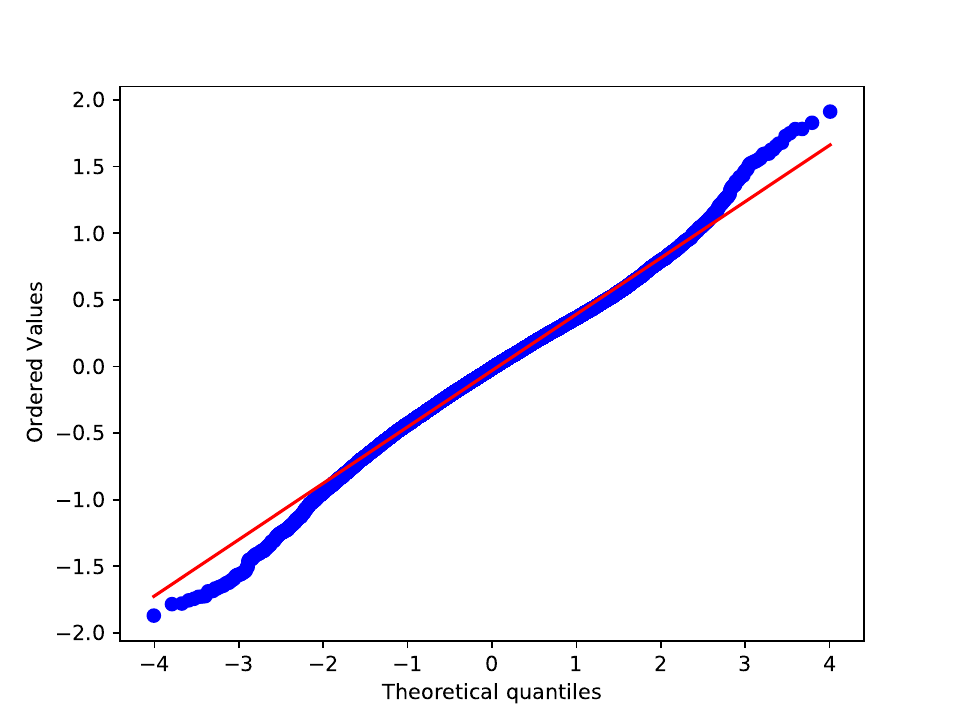}
    }
    \subfigure[d = 507]{
        \centering
        \includegraphics[width=0.18\linewidth]{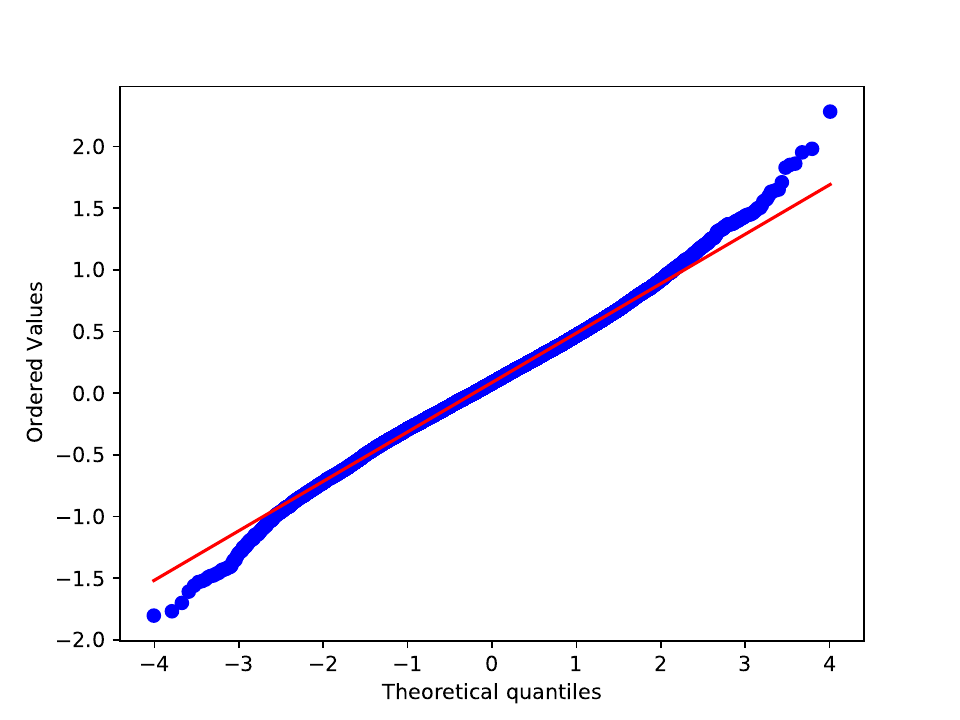}
    }
    \subfigure[d = 527]{
        \centering
        \includegraphics[width=0.18\linewidth]{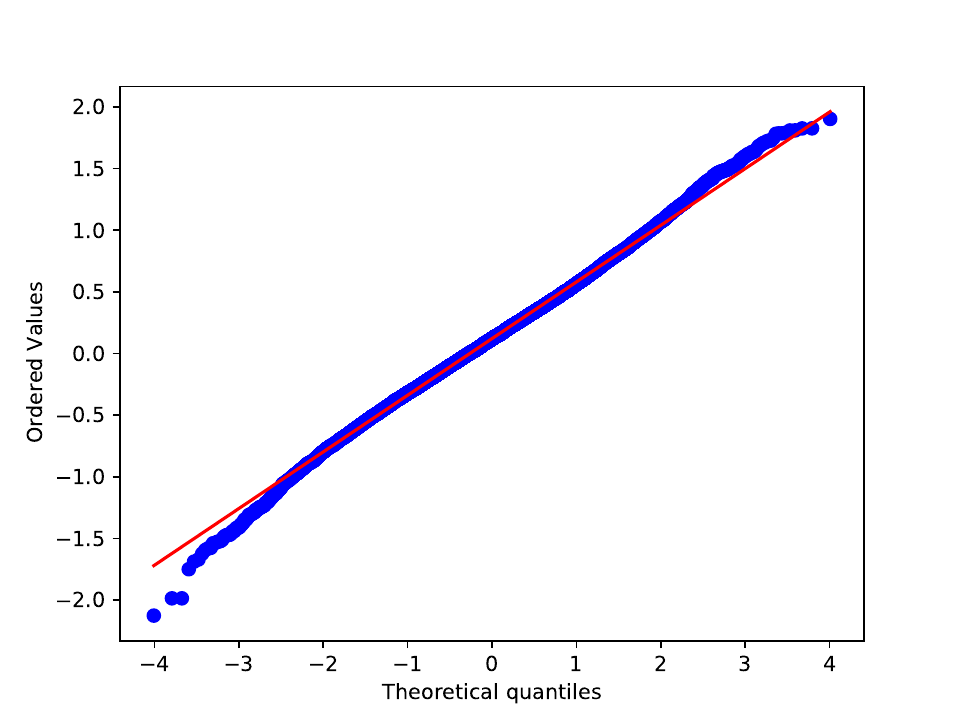}
    }
    \subfigure[d = 655]{
        \centering
        \includegraphics[width=0.18\linewidth]{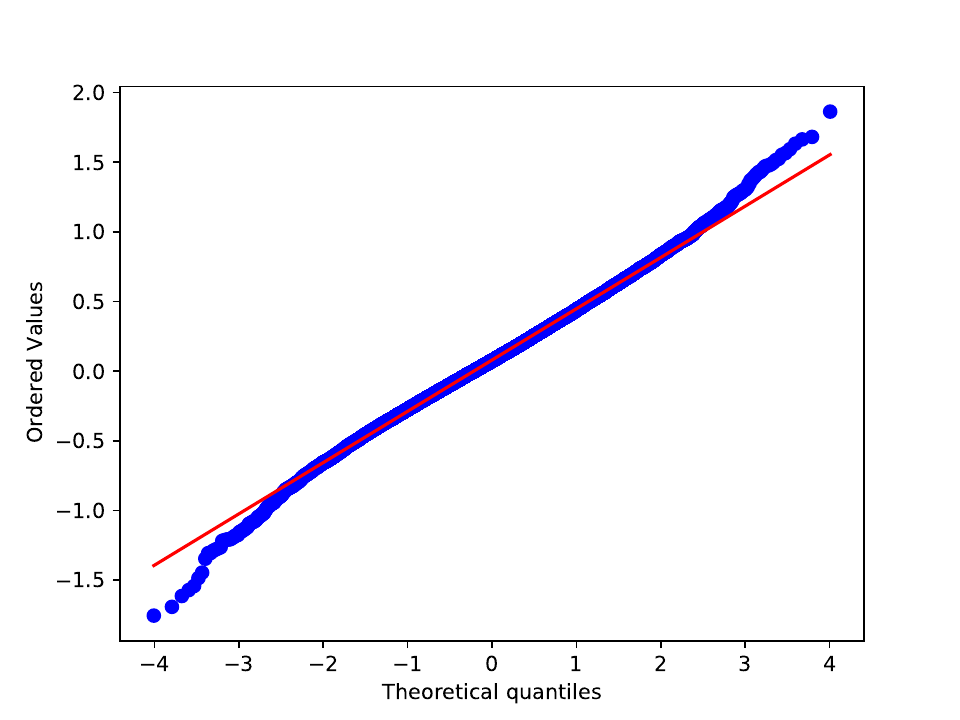}
    }
    \subfigure[d = 670]{
        \centering
        \includegraphics[width=0.18\linewidth]{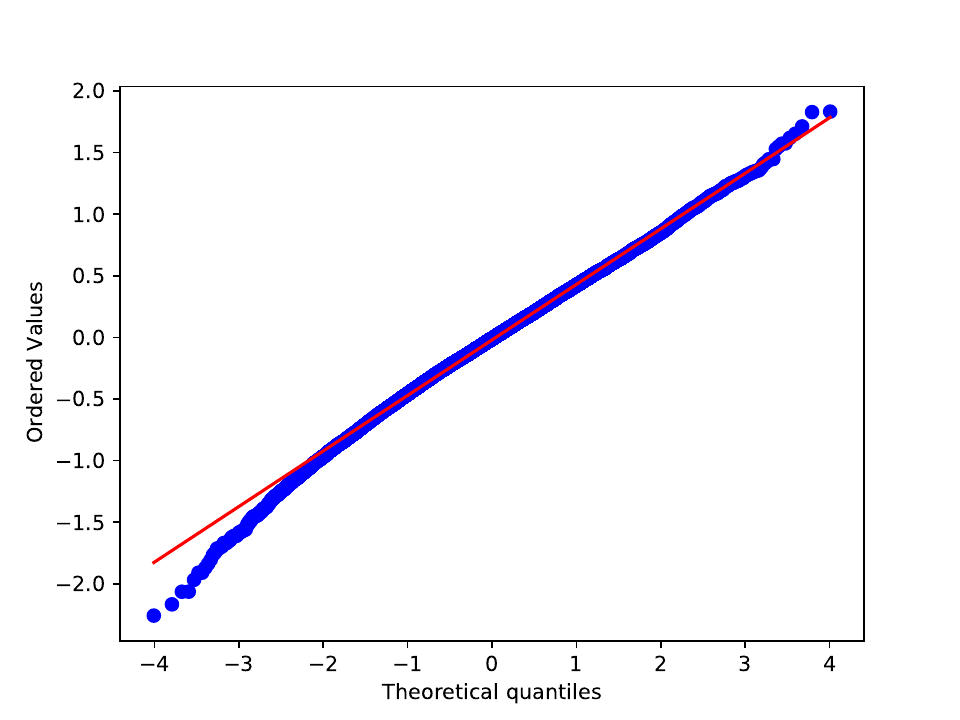}
    }
    
    \vspace{-3mm}
    \caption{Q-Q plots on ten random dimensions.}
    \label{fig:qq10}
    \vspace{-3mm}
\end{figure*}

We randomly sample another 10 dimensions from the 768 dimensions of BERT and plot the quantiles against a normal distribution in Figure~\ref{fig:qq10}. All the 10 dimensions match with a normal distribution pretty well. 

\begin{table}[h]
\centering
\caption{Representations following $d$ normal distributions with parameters estimated from neural language models.}
%  \jingbo{this caption could be improved.}
\vspace{-3mm}
\label{tbl:normal_dist_appendix}
\small
\begin{tabular}{l c c} 
\toprule
Model & Average & Min\\
\midrule
BERT & .9995 & .9978\\
RoBERTa & .9989 & .9982 \\
GPT-2 & .9988 & .9982\\
XLNet & .9994 & .9977 \\
ELMo & .9987 & .9947\\
\bottomrule
\end{tabular}
\vspace{-3mm}
\end{table}
\section{Normal Distribution Estimated from Models}
In additional to randomly sampled $\mu_j$ and $\sigma_j$, we can also use the empirical mean and standard deviation of (dimensions of) representations from pre-trained language models. Table~\ref{tbl:normal_dist_appendix} shows that the property is well satisfied on these representations. This further advocates that representations from these models have properties similar to normal distributions.

\section{Diagonal \& Off-diagonal Values}
Here we show the calculations for values in the covariance matrix $C$.
Note that

$$
\begin{aligned}
C_{ij} = \frac{1}{d - 1} \sum_{k = 1}^d r_{ki}r_{kj},
\end{aligned}
$$ 
so for diagonal entries $C_{ii}$ is a sum of $d$ products of normally distributed random variables with itself, and all $C_{ii}$ follow the same distribution; for off diagonal entries $C_{ij}$ is a sum of $d$ products of pairs of normally distributed random variables, and similarly, all off diagonal entries also follow the same distribution. Therefore, on expectation, the covariance matrix have the same diagonal entries, and the same off-diagonal entries. The average and variance can be mathematically derived:

\vspace{-3mm}
\begin{equation*}
    \scriptsize
    \begin{split}
    \scriptsize
        \mathbb{E}[C_{ii}] & = \frac{1}{d - 1} \sum_{k = 1}^d (\sigma_k^2 + \mu_k^2) \\
Var[C_{ii}] & = \frac{1}{(d - 1)^2} \left(\sum_{k = 1}^d 2\sigma_k^4 + 4\mu_k^2\sigma_k^2\right) \\
\mathbb{E}[C_{ij}] & = \frac{1}{d - 1} \sum_{k = 1}^d \mu_k^2 \\
Var[C_{ij}] & = \frac{1}{(d - 1)^2} \left(\sum_{k = 1}^d \sigma_k^4 + 2\mu_k^2\sigma_k^2\right)
    \end{split}
\end{equation*}

We also outline steps for the derivation. Following our notations, $r_{ij} \sim N(\mu_j, \sigma_j^2) \implies r_{ij} = \sigma_j * z_{ij} + \mu_j$ where $z_{ij}$ is a standard normal variable, i.e. $z_{ij} \sim N(0, 1)$.
 
\begin{figure*}
\centering
\begin{equation}
    \begin{split}
        E[C_{ii}] &= E[\frac{1}{d - 1} * \sum_{k = 1}^d r_{ik}r_{ik}] \\
        &= \frac{1}{d - 1} \sum_{k = 1}^d E[(\sigma_k * z_{ik} + \mu_k)^2] \\
        &= \frac{1}{d - 1} \sum_{k = 1}^d (\sigma_k^2 + \mu_k^2)
    \end{split}
\end{equation}
\end{figure*}

\begin{figure*}
\centering
\begin{equation}
    \begin{split}
        E[C_{ij}] &= E[\frac{1}{d - 1} * \sum_{k = 1}^d r_{ik}r_{jk}] \\
        &= \frac{1}{d - 1} \sum_{k = 1}^d E[(\sigma_k * z_{ik} + \mu_k)(\sigma_k * z_{jk} + \mu_k)] \\
        &= \frac{1}{d - 1} \sum_{k = 1}^d (\mu_k^2)
    \end{split}
\end{equation}
\end{figure*}

\begin{figure*}
\centering
\begin{equation}
    \begin{split}
        Var[C_{ii}] &= E[\left(\frac{1}{d - 1} * \sum_{k = 1}^d r_{ki}r_{ki} \right)^2] - E[C_{ii}]^2 \\
        &= \frac{1}{(d - 1)^2} E[(\sum_{k=1}^d r_{ki} * r_{ki})^2] - E[C_{ii}]^2
    \end{split}
\end{equation}

\begin{equation}
    \begin{split}
        E[(\sum_{k=1}^d r_{ki} * r_{ki})^2] &= E[(\sum_{k=1}^d (\sigma_k z_{ik} + \mu_k)^2)^2]\\
        &= \sum_{k=1}^d E[(\sigma_k^2z_{ik}^2 + 2\mu_k\sigma_kz_{ik} + \mu_k^2)^2] \\ 
        & \textrm{\;\;\;\;} + \sum_{k_1 != k_2} E[(\sigma_{k_1}^2z_{i{k_1}}^2 + 2\mu_{k_1}\sigma_{k_1}z_{i{k_1}} + \mu_{k_1}^2) * (\sigma_{k_2}^2z_{i{k_2}}^2 + 2\mu_{k_2}\sigma_{k_2}z_{i{k_2}} + \mu_{k_2}^2) ] \\
        &= \sum_{k=1}^d E[(\sigma_k^2z_{ik}^2 +\mu_{k}^2)^2 + 4\mu_k^2\sigma_k^2z_{ik}^2] + \sum_{k_1 != k_2} (\sigma_{k_1}^2 + \mu_{k_1}^2) * (\sigma_{k_2}^2 + \mu_{k_2}^2) \\
        &= \sum_{k=1}^d \sigma_k^4 * 3 + \mu_{k}^4 + 2 * \sigma_k^2\mu_k^2 + 4\mu_k^2\sigma_k^2 + \sum_{k_1 != k_2} (\sigma_{k_1}^2 + \mu_{k_1}^2) * (\sigma_{k_2}^2 + \mu_{k_2}^2) \\
        &= \left(\sum_{k=1}^d \sigma_{k}^2 + \mu_{k}^2 \right)^2 + \sum_{k = 1}^d 2\sigma_k^4 + 4\mu_k^2\sigma_k^2
    \end{split}
\end{equation}
\end{figure*}

\begin{figure*}
\centering
\begin{equation}
    \begin{split}
        Var[C_{ij}] &= E[\left(\frac{1}{d - 1} * \sum_{k = 1}^d r_{ki}r_{kj} \right)^2] - E[C_{ij}]^2 \\
        &= \frac{1}{(d - 1)^2} E[(\sum_{k=1}^d r_{ki} * r_{kj})^2] - E[C_{ij}]^2
    \end{split}
\end{equation}
\end{figure*}
\begin{figure*}
\centering
\begin{equation}
    \begin{split}
        E[(\sum_{k=1}^d r_{ki} * r_{kj})^2] &= E[\left(\sum_{k=1}^d (\sigma_k z_{ik} + \mu_k) * (\sigma_k z_{jk} + \mu_k)\right)^2]\\
        &= \sum_{k=1}^d E[(\sigma_k^2z_{ik}z_{jk} + \mu_k\sigma_k(z_{ik} + z_{jk}) + \mu_k^2)^2] \\ 
        & \;\; + \sum_{k_1 != k_2} E[(\sigma_{k_1}^2z_{i{k_1}}z_{j{k_1}} + \mu_{k_1}\sigma_{k_1}(z_{i{k_1}}+z_{j{k_1}}) + \mu_{k_1}^2) \\
        & \;\; * (\sigma_{k_2}^2z_{i{k_2}}z_{j{k_2}} + \mu_{k_2}\sigma_{k_2}(z_{i{k_2}}+z_{j{k_2}}) + \mu_{k_2}^2) ] \\
        &= \sum_{k=1}^d E[(\sigma_k^2z_{ik}z_{jk} +\mu_{k}^2)^2 + \mu_k^2\sigma_k^2(z_{ik} + z_{jk})^2] + \sum_{k_1 != k_2} \mu_{k_1}^2 * \mu_{k_2}^2 \\
        &= \sum_{k=1}^d \sigma_k^4 + \mu_{k}^4 + \mu_k^2\sigma_k^2 * 2 + \sum_{k_1 != k_2} \mu_{k_1}^2 * \mu_{k_2}^2 \\
        &= \left(\sum_{k=1}^d \mu_{k}^2 \right)^2 + \sum_{k = 1}^d \sigma_k^4 + 2\mu_k^2\sigma_k^2
    \end{split}
\end{equation}
\end{figure*}

\end{document}